# ICDAR 2019 Robust Reading Challenge on Reading Chinese Text on Signboard

Xi Liu[1], Rui Zhang[1], Yongsheng Zhou[1], Qianyi Jiang[1], Qi Song[1], Nan Li[1], Kai Zhou[1], Lei Wang[1], Dong Wang[1],
Minghui Liao[2], Mingkun Yang[2], Xiang Bai[2], Baoguang Shi[3], Dimosthenis Karatzas[4], Shijian Lu[5], C. V. Jawahar[6]

[1]Meituan-Dianping Group, China, [2]School of EIC, Huazhong University of Science and Technology, China,
[3]Microsoft Redmond, USA, [4]Computer Vision Centre, UAB, Spain, [5]Nanyang Technological University, Singapore, [6]IIIT Hyderabad, India

*Abstract*—Chinese scene text reading is one of the most challenging problems in computer vision and has attracted great interest. Different from English text, Chinese has more than 6000 commonly used characters and Chinese characters can be arranged in various layouts with numerous fonts. The Chinese signboards in street view are a good choice for Chinese scene text images since they have different backgrounds, fonts and layouts. We organized a competition called ICDAR2019-ReCTS, which mainly focuses on reading Chinese text on signboard. This report presents the final results of the competition. A large-scale dataset of 25,000 annotated signboard images, in which all the text lines and characters are annotated with locations and transcriptions, were released. Four tasks, namely character recognition, text line recognition, text line detection and end-to-end recognition were set up. Besides, considering the Chinese text ambiguity issue, we proposed a multi ground truth (multi-GT) evaluation method to make evaluation fairer. The competition started on March 1, 2019 and ended on April 30, 2019. 262 submissions from 46 teams are received. Most of the participants come from universities, research institutes, and tech companies in China. There are also some participants from the United States, Australia, Singapore, and Korea. 21 teams submit results for Task 1, 23 teams submit results for Task 2, 24 teams submit results for Task 3, and 13 teams submit results for Task 4. The official website for the competition is http://rrc.cvc.uab.es/?ch=12.

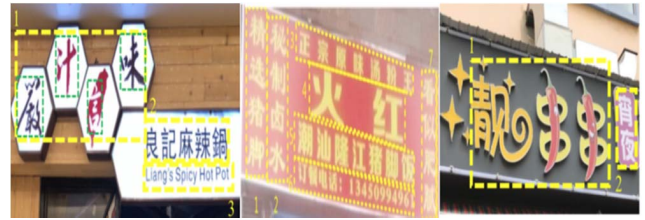

Figure 1. Characters with various layouts and fonts.

## I. INTRODUCTION

Texts in natural images carry much important semantic information. Reading text in natural scene images has been widely studied recently since it is an important prerequisite for many content-based image analysis tasks such as photo translation, fine-grained image classification and autonomous driving.

It is widely recognized that large-scale, well-annotated datasets are crucial to today's deep learning based techniques. In scene text reading field, many scene text datasets have been collected. Especially for Chinese text reading, more and more Chinese scene text datasets are proposed, such as MSRA-500 [1], RCTW [2], SCUT-CTW1500 [3], CTW [4].

Chinese text reading is a huge challenge task. Different from English text reading, Chinese has more than 6000 commonly used characters. Besides, owing to the Chinese culture, the layouts, arrangements and fonts of Chinese characters are always in a great variety, as shown in Figure 1.

The Chinese signboards in street view may be the best source for Chinese scene text images since they have different backgrounds, fonts and layouts. In Meituan-Dianping Group, a Chinese leading company for food delivery services, consumer products and retail services, there are many signboard images collected by Meituan business merchants. Based on this, we propose a competition for Chinese text reading on signboard and construct a large-scale challenging natural scene text dataset of 25,000 signboard images. About 200,000 text lines and 600,000 characters are labeled with locations and transcriptions. We set up four tasks for this competition, namely character recognition, text line recognition, text line detection and end-to-end recognition. Besides, we propose a multi ground truth (multi-GT) evaluation method considering the Chinese text ambiguity. As illustrated in Figure 2, it is difficult to determine whether some words should be merged to a text instance or not. We thus provide one or more ground truths for each test image and compare the predicted result with all the ground truths when evaluating. The best matched GT will be used to calculate the evaluation metrics.

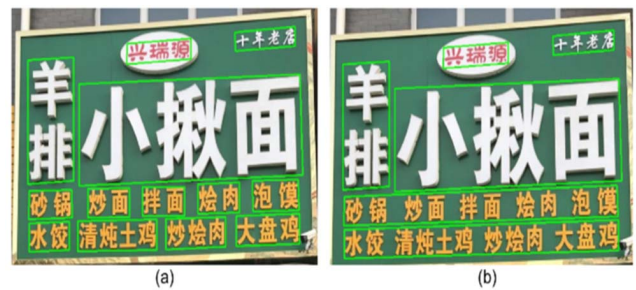

Figure 2. Chinese text ambiguity in signboard image.

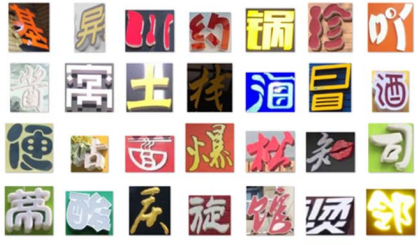

Figure 3. Chinese character test images.

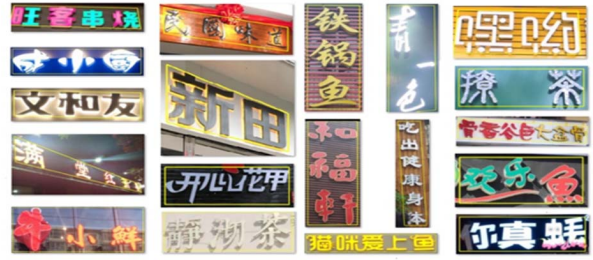

Figure 4. Text line test images.

The competition lasts from March 1st to April 30, 2019. It receives lots of attention from the community. For all the four tasks, there are all together 46 valid teams participating in the competition and hundreds of valid submissions are received. In this report, we will present their evaluation results.

## II. DATASET AND ANNOTATIONS

The dataset, named ReCTS-25k, comprises 25,000 signboard images. All the images are from Meituan-Dianping Group, collected by Meituan business merchants, using phone cameras under uncontrolled conditions. Different from other datasets, this dataset mainly focuses on Chinese text reading on the signboards. The layout and arrangement of Chinese characters in signboards are much more complex for the sake of aesthetics appearance or highlighting certain elements. Figure 1 shows some example images.

We manually annotate the locations and transcriptions for all the text lines and characters in the signboard images. Note that the utterly obscure and small text lines and characters are marked with a difficult flag. Locations are annotated in terms of polygons with four vertices, which are in clockwise order starting from the upper left vertex. Transcriptions are UTF-8 encoded strings.

The dataset is split into two subsets. The training set consists of 20,000 images, and the test set consists of 5,000 images. Moreover, 29335 character images and 10789 text lines images, cropped from the 5000 test images, are used for task 1 and task 2 evaluation respectively.

## III. CHALLENGE TASKS

Robust reading challenge on Chinese signboard consists of four tasks: 1) Character recognition, 2) Text line recognition, 3) Text line detection, 4) End-to-end recognition. Given that Chinese signboards have various layouts, fonts and orientations, character and text line reading are concerned. Therefore, in our competition, character based and text line based tasks are both evaluated.

Note that the half-width character and its corresponding full-width character are regarded as one character in the evaluation of task 2 and task 4. Moreover, the English letters are not case sensitive.

### A. Task 1 – Character Recognition

The aim of this task is to recognize characters of the cropped character images from Chinese signboards. As illustrated in Figure 3, the Chinese characters take the largest portion and are in diverse fonts. Participant is asked to submit a text file containing character results for all test images. The recognition accuracy is given as the metric:

$$\text{accuracy} = \frac{N_{right}}{N_{total}}, \quad (1)$$

where $N_{right}$ is the number of characters predicted correctly and $N_{total}$ is the total number of the test characters.

### B. Task 2 – Text Line Recognition

The target of text line recognition is to recognize the cropped word images of scene text. The cropped text line images as well as the coordinates of the polygon bounding boxes in the images are given. The given points are arranged in the clockwise order, starting from the top-left point. Figure 4 shows some examples of the test set. The text line images may contain perspective and arbitrary arranged text lines.

The results are evaluated by the Normalized Edit Distance between the recognition result and the ground truth. The edit distances are summarized and divided by the number of test images. The resulting average edit distance is taken as the metric and is formulated as follows:

$$\text{accuracy} = 1 - \frac{1}{N}\sum_{i=1}^{N}\frac{D(s_i,\hat{s}_i)}{\max(s_i,\hat{s}_i)}, \quad (2)$$

where $D$ stands for the Levenshtein Distance, $s_i$ denotes the predicted text line and $\hat{s}_i$ denotes the corresponding ground truth, $N$ is the total number of text lines.

### C. Task 3 – Text Line Detection

The aim of this task is to localize text lines in the signboard. The input image is the full signboard images. The detection results submitted by the participants are required to give four vertices of the polygon in clockwise order.

In some signboard, there always exist the following case, as shown in Figure 2. It is difficult to determine whether the boxes "砂锅"，"炒面"，"拌面"，"烩肉"，"泡馍" should be merged to a large text box or not. Therefore, we regard the two cases (Figure 2(a) and Figure 2(b)) as correct ground truth. We provide one or more ground truths for each test image. When evaluating, we compare the predicted result with all the ground truths and use the best matched one to calculate the evaluation metrics.

Following the evaluation protocols of ICDAR 2017-RCTW [2] dataset, the detection task is evaluated in terms of Precision, Recall and F-score with intersection-over-union (IoU) threshold of 0.5 and 0.7. The F-score at IoU=0.5 will be used as the only metric for the final ranking. All detected or missed ignored ground truths will not contribute to the evaluation result.

*D. Task 4 – End-to-End Recognition*

The aim of this task is to localize and recognize every text instance in the signboard. The input image is the full signboard images. Participants are required to submit the text file containing all the recognized text lines locations and transcriptions for each test image. Similar to Task 3, the locations are four vertices in clock-wise order and the transcripts are UTF-8 encoded strings.

The evaluation process consists of two steps. First, each detection is matched to a ground truth polygon that has the maximum IOU, or it is matched to 'None' if none IOU is larger than 0.5. If multiple detections are matched to the same ground-truth, only the one with the maximum IOU will be kept and the others are recorded as 'None'. Then, we calculate the edit distances between all matching pairs by Formula (2). Since one test image may have multiple ground truths, as stated in Task 3, we also compare the predicted result with all the ground truths and use the best matched one to calculate the evaluation metrics.

## IV. ORGANIZATION

The competition starts on March 1, 2019, when the RRC website is ready and open for registration. The training set is released on March 18, the first part of test set is released on April 12 and the second part of test set released on April 20. We revise the test set more than once to fixed some errors before releasing the test set. The RRC website opens for result submission on April 20 and closes at 11:59 PM PST, April 30.

There are all together 46 valid teams participated in the competition. Most of the participants come from universities, research institutes, and tech companies in China. There are also some participants from the United States, Australia, Singapore, and Korea.

All the teams submit their results through the RRC website. Each team is allowed to submit 5 results at most and we choose the best result among the 5 results as the final result. 21 teams submit results for Task 1, 23 teams submit results for Task 2, 24 teams submit results for Task 3, and 13 teams submit results for Task 4.

## V. SUBMISSIONS AND RESULTS

The evaluation script is implemented in Python. We run the script to evaluate all the submissions. Table I summarizes the top 5 results of Task 1. Methods are ranked by their accuracy. Table II summarizes the top 5 results of Task 2. Methods are ranked by their normalized edit distance. Table III summarizes the top 5 results of Task 3. Methods are ranked by their F-score.

Table IV summarizes the top 5 results of Task 4. Methods are ranked by their normalized edit distance. You can view the complete ranking in the home page of the competition https://rrc.cvc.uab.es/?ch=12.

*A. Top 3 submissions for Task 1*

**1. "BASELINE v1" (USTC-iFLYTEK)** The method uses image classification methods and its ensemble.

**2. "Amap_CVLab" (Alibaba AMAP)** The method adds res-block [5] (for the lower dimension feature collapse avoiding) and se-block [6]. Their training dataset contains both the ReCTS-25k and other data.

**3. "TPS-ResNet v1" (Clova AI OCR Team, NAVER/LINE Corp)** The method uses Thin-plate-spline(TPS) [7] based Spatial transformer network (STN) [8], which normalizes the input text images. They use ResNet [5], BiLSTM [9] and attention mechanism. Their training dataset contains the Chinese synthetic datasets (MJSynth and SynthText [10]) and real dataset (ArT [11], LSVT [12], RCTW [2], ReCTS-25k).

*B. Top 3 submissions for Task 2*

**1. "SANHL" (South China University of Technology, Northwestern Polytechnical University, The University of Adelaide, Lenovo and Huawei)** The method uses an ensemble framework, which consists of attention-based network, transformer network and CTC-based [13] network. Apart from the official training dataset, about 2 million synthesized samples are used for training.

**2. "Tencent-DPPR Team" (Tencent-DPPR Team)** The method uses five types of deep models, which mainly include CTC-based nets and multi-head attention based nets. All samples are resized to the same height before feeding into the network. Furthermore, besides ReCTS, they use a synthetic dataset containing more than fifty million images, as well as open-source datasets including LSVT [12], COCO-Text [14], RCTW [2] and ICPR-2018-MTWI. In terms of data augmentation, they mainly use Gaussian blur, Gaussian noise and so on.

**3. "HUST_VLRGROUP" (Huazhong University of Science and Technology)** A CRNN based method.

*C. Top 3 submissions for Task 3*

**1. "SANHL_v4" (South China University of Technology, The University of Adelaide, Northwestern Polytechnical University, Lenovo, HUAWEI)** The method uses a sequential-free box discretization method to localize the text instances. Multi-scale testing and model ensemble are used to generate the final result. Their training dataset contains LSVT [12], ArT [11], MLT [15] and ReCTS-25k.

**2. "Tencent-DPPR Team" (Tencent Data Platform Precision Recommendation)** Their text detector is based on two-stage method with multi-scale training policy, and ResNet101 [5] is used as the backbone network. They use feature pyramid layers [16] to extract features instead of choosing one layer according to box sizes. They use LSVT [12] pre-trained model.

**3. "Amap-CVLab" (Alibaba AMAP, Alibaba DAMO Academy for Discovery, Adventure, Momentum and Outlook)** The method is based on Mask R-CNN [17]. Their training dataset contains RCTW[2], ICDAR2017-MLT[15], LSVT[12], ReCTS-25k.

*D. Top 3 submissions for Task 4*

**1. "Tencent-DPPR Team" (Tencent-DPPR Team)** In the detection part, they use a text detector based on two-stage method. This method uses ResNet101 [5] as feature extractor, and they design a policy to help proposals select feature pyramid layers [16] to extract features instead of choosing one layer according to box sizes. In detection ensemble stage, they apply a multi-scale test method with different backbones. When ensembling all the results, they develop an approach to vote boxes after scoring each box. In the recognition part, they use an ensemble model, which includes CTC-based nets and multi-head attention based nets. For this task, they use the predicted confidence scores of cropped words and the ensemble results to select the reliable one among results predicted by all models.

**2. "SANHL" (South China University of Technology, Northwestern Polytechnical University, The University of Adelaide, Lenovo and Huawei)** The method firstly detect possible text lines, and then predict strings by an ensembled recognition model.

**3. "HUST_VLRGROUP" (Huazhong University of Science and Technology)** The method uses Mask R-CNN as text detector and a CRNN based approach to predict strings.

*E. Baseline submissions*

For reference, we submit a baseline method to Task 1, Task 2, Task 3 and Task 4 respectively. The methods are implemented by ourselves. Their results are shown in Table I, II, III and IV.

For Task 1, the character Recognition method is based on the densely connected convolutional network (DenseNet) [18]. Our network inherits from the DenseNet-169 network model with dense blocks, but we reduce the number of last dense block to 24 and all the growth rates in the networks are 32. The training dataset consists of ReCTS and synthetic data.

For Task 2, We took the Chinese text line recognition as a sequence recognition task. We utilized a modified version of Inception-V4 [19], integrated with attention module to extract feature maps. The CTC layer for transcription is adopted. The baseline result is obtained by a single recognition model, the training dataset consists of ReCTS, RCTW [2], and LSVT [12], no synthetic data is utilized.

For Task 3, the text detection method is based on SEG-FPN [20] and Pixel-link [21]. We build a unified framework, which combines pixel link and segment link in feature pyramid network to detect scene text. The training dataset only consists of ReCTS.

For Task 4, we first detect the text line in the image. If the text line is horizontal, recognize it by the line recognition model; if the text line is vertical, character detection and character recognition model will be used. The text line detection part is the same as that for Task 3, the character recognition part is the same as that for Task 1, and the text line recognition part is the same as that for Task 2. A Faster-RCNN [22] based detection approach is adopted to detect Chinese character regions.

VI. CONCLUSIONS

We organize the competition on reading Chinese text on signboard (ReCTS). A large-scale challenging natural scene text dataset of 25,000 signboard images are released and four tasks are set up. We also propose a multi-GT evaluation strategy intended for Chinese text ambiguity. During the challenge, we receive hundreds of submissions from 46 teams, which shows the broad interest in the community. In the future, we plan to make the evaluation scripts available on the website https://rrc.cvc.uab.es/ and users can get the evaluation results shortly after they submit the results to the website.

TABLE I: RESULTS SUMMARY FOR THE TOP-5 SUBMISSIONS OF TASK 1.

| Ranking | Team Name | Affiliation | Accuracy |
|---|---|---|---|
| 1 | BASELINE-v1 | iFLYTEK, University of Science and Technology of China | 0.9737 |
| 2 | Amap_CVLab | Alibaba AMAP | 0.9728 |
| 3 | TPS-ResNet-v1 | Clova AI OCR Team, NAVER/LINE Corp | 0.9612 |
| 4 | SANHL_v4 | South China University of Technology, The University of Adelaide, Northwestern Polytechnical University, Lenovo, HUAWEI | 0.9594 |
| 5 | Tencent-DPPR | Tencent (Data Platform Precision Recommendation) | 0.9512 |
| **Baseline** | | **Meituan Dianping** | **0.9140** |

TABLE II: RESULTS SUMMARY FOR THE TOP-5 SUBMISSIONS OF TASK 2.

| Ranking | Team Name | Affiliation | N.E.D |
|---|---|---|---|
| 1 | SANHL_v1 | South China University of Technology, The University of Adelaide, Northwestern Polytechnical University, Lenovo, HUAWEI | 0.9555 |
| 2 | Tencent-DPPR | Tencent (Data Platform Precision Recommendation) | 0.9486 |
| 3 | HH-Lab-v4 * | Huazhong University of Science and Technology (Visual and Learning Representation Group) | 0.9483 |
| 4 | TPS-ResNet-v1 | Clova AI OCR Team, NAVER/LINE Corp | 0.9477 |
| 5 | Baseline-Beihang* | Beihang University | 0.9437 |
| **Baseline** | | **Meituan Dianping** | **0.9089** |

TABLE III: RESULTS SUMMARY FOR THE TOP-5 SUBMISSIONS OF TASK 3.

| Ranking | Team Name | Affiliation | F-score |
|---|---|---|---|
| 1 | SANHL_v4 | South China University of Technology, The University of Adelaide, Northwestern Polytechnical University, Lenovo, HUAWEI | 0.9336 |
| 2 | Tencent-DPPR | Tencent (Data Platform Precision Recommendation) | 0.9303 |
| 3 | Amap-CVLab | Alibaba AMAP, Alibaba DAMO Academy for Discovery, Adventure, Momentum and Outlook | 0.9250 |
| 4 | HH-Lab * | Huazhong University of Science and Technology (Visual and Learning Representation Group) | 0.9127 |
| 5 | maskrcnn_text * | Huazhong University of Science and Technology (Media and Communication Laboratory, Text detection) | 0.9102 |
| **Baseline** | | **Meituan Dianping** | **0.9001** |

TABLE IV: RESULTS SUMMARY FOR THE TOP-5 SUBMISSIONS OF TASK 4.

| Ranking | Team Name | Affiliation | N.E.D |
|---|---|---|---|
| 1 | Tencent-DPPR | Tencent (Data Platform Precision Recommendation) | 0.8150 |
| 2 | SANHL_v1 | South China University of Technology, The University of Adelaide, Northwestern Polytechnical University, Lenovo, HUAWEI | 0.8144 |
| 3 | HH-Lab * | Huazhong University of Science and Technology (Visual and Learning Representation Group) | 0.7943 |
| 4 | baseline_Beihang * | Beihang University | 0.7661 |
| 5 | SECAI * | Institute of Information Engineering, Chinese Academy of Sciences, University of Science & Technology Beijing | 0.7437 |
| **Baseline** | | **Meituan Dianping** | **0.7298** |

* means student contestant